\documentclass{article}

\usepackage[preprint]{neurips_2026}
\usepackage{amsmath}
\usepackage{amssymb}
\usepackage{algorithm}
\usepackage{algorithmic}
\usepackage{microtype}
\usepackage{subcaption}
\usepackage{booktabs} 
\usepackage{graphicx} 
\usepackage[table]{xcolor}
\usepackage[export]{adjustbox} 
\usepackage{amssymb}
\usepackage{mathtools}
\usepackage{amsthm}
\theoremstyle{plain}

\theoremstyle{definition}

\theoremstyle{remark}

\usepackage{hyperref}
\usepackage{cleveref}
\usepackage{xcolor}
\usepackage{colortbl}
\usepackage{booktabs}
\usepackage{array}

\usepackage{xcolor}
\usepackage{colortbl}
\usepackage{booktabs}

\definecolor{bestcolor}{RGB}{31,78,121}
\definecolor{secondcolor}{RGB}{85,95,105}
\definecolor{rowgray}{RGB}{235,238,242}

\newcommand{\best}[1]{\textbf{\textcolor{bestcolor}{#1}}}
\newcommand{\second}[1]{\textcolor{secondcolor}{\underline{#1}}}

\usepackage[utf8]{inputenc} 
\usepackage[T1]{fontenc}    
\usepackage{hyperref}       
\usepackage{url}            
\usepackage{booktabs}       
\usepackage{amsfonts}       
\usepackage{nicefrac}       
\usepackage{microtype}      
\usepackage{xcolor}         
\usepackage{wrapfig}
\title{Spokes: Optimizing for Diverse Pretraining Data Selection}

\author{%
  Clarence Lee \\
  DSO National Laboratories\\
  \And
  Yejin Choi \\
  Stanford University
  \And
  Luke Zettlemoyer \\
  University of Washington \\
  \And
  Pang Wei Koh \\
  University of Washington \\
  \And
  Hai Leong Chieu \\
  DSO National Laboratories \\
}

\begin{document}
\maketitle
\begin{abstract}
Diversity plays a critical role in data selection, improving performance under fixed data budgets by reducing redundancy and repetition. However, optimizing for diversity is inherently challenging, as it is a set-level property that depends on interactions between data points rather than individual examples. As a result, existing approaches typically rely on proxies or approximations, which often fail to ensure sufficiently diverse subsets. In this work, we directly optimize diversity by introducing a probabilistic diversification framework based on the G-Vendi score, optimized via exponentiated gradient descent. Our method produces subsets that are substantially more diverse than those obtained via random sampling, achieving a +489 increase in G-Vendi score on a 500k-sample subset. We evaluate our approach on FineWeb and DCLM, where it consistently outperforms existing methods. Notably, \textsc{Spokes} (diversity-only) improves average downstream performance by +0.4 and +0.5 points over random sampling on DCLM and FineWeb, respectively. More importantly, jointly optimizing for both quality and diversity yields the strongest results: \textsc{Spokes} achieves gains of +1.5 and +1.4 points on DCLM and FineWeb, outperforming all baselines, including semantic deduplication and quality filtering.

\end{abstract}

\section{Introduction}

Data diversity is a key factor in the construction of pretraining corpora. Prior work has shown that explicitly incorporating diversity into data mixture design and optimization can improve downstream performance \citep{liu2025quadmix, fan2025joint, jung2025prismatic}. However, diversity is fundamentally a set-level objective, which means that it is measured as a property of a set rather than an individual element, and directly optimizing it is computationally intractable due to the combinatorial nature of subset selection, an issue that becomes more pronounced at pretraining scale.
To address this, practical methods have been developed and are already used in modern pretraining pipelines. These approaches typically rely on coarse grained structure, such as clustering data into topics \citep{wettig2025organize}, skills \citep{chandiramani2026nemotron}, or unsupervised clustering \citep{liu2025quadmix, diao2025nemotron}. While these methods are effective at approximating diversity by ensuring each cluster is represented in the final mix, they still rely on low resolution partitions of data, and effective diversity can at best be approximated by random sampling of data. As a result, information over long-tail or underrepresented knowledge may not be adequately captured. The use of finegrained signals from individual data points can be a more reliable proxy to build highly diverse sets, but a gap still remains in literature on how to scale this reliably.

This gap motivates our central question:
\textit{How may we reliably extract diverse subsets from pretraining corpora at scale, in order to lead to better downstream outcomes?}

Just like how diverse normalised vectors distribute evenly on a unit circle in 2d space, which capture the imagery of spokes on a wheel, our method,  \textsc{Spokes} is a principled approach to obtain diverse sets of data. Rather than rely on heuristics, we solve a global optimization that analyzes the contribution of every data point. We are then able to use this to extract data points that contribute to the overall diversity of the set. 

Our method reveals that dense, highly diverse subsets can be extracted from pretraining corpora. While the diversity scores of random samples saturates quickly, selecting data points using the optimised weights from \textsc{Spokes} led to a siginificant 489 point increase in G-Vendi \citep{jung2025prismatic} scores on a 500k sample subset. This is in contrast to existing methods that try to improve diversity such as SemDeDup \citep{abbas2023semdedup}, which only led to a 7 point increase in G-Vendi scores. 

While existing measures treat the G-Vendi score as a post-hoc evaluation measure for diversity, we directly optimise for it and introduce a set of practical strategies (See \Cref{sec:scaling_spokes}) that led us to successfully extract diverse sets from large scale pretraining corpora from Fineweb and DCLM. Not only do batch level diversity scores increase with the chosen subsets, we demonstrate its success in improving downstream performance on both datasets. This shows that the benefits of diversity reliably translate across different datasets which contain different levels of filtering and quality.

While quality filters have traditionally been a strong baseline for data selection, as many quality signals are designed to correlate with evaluation performance, \textsc{Spokes} demonstrates that diversity is also an important axis to consider. By balancing quality scores, which were based on model based classifiers (2.95 vs 3.18), for batches with higher G-vendi scores (425 vs 315). We were able to achieve significant improvements in evaluation score over the quality only baseline, achieving +1.0 and + 1.9 in DCLM and Fineweb respectively.

As such, we investigate direct diversity optimization for data selection at pretraining scale. Our contributions are as follows: (1) We introduce \textsc{Spokes}, a scalable techniques for diversity
optimization that operate efficiently at the scale of modern pretraining datasets. (2) We show that \textsc{Spokes} is effective, and is able to extract highly diverse subsets from existing pretraining corpora.
(3) We jointly optimize for both quality and diversity which yields consistent gains in pretraining
performance.

\section{Background}
\subsection{G-Vendi as a diversity measure}
The \emph{Vendi} score \citep{friedman2022vendi} was introduced as a principled metric for quantifying the diversity of a dataset.  Concretely, given a set of representations, the Vendi score constructs a similarity matrix and examines the exponential of the entropy of its' spectrum (eigenvalues). When the data points are very similar, most of the spectrum is concentrated in a few directions. When the data points are diverse, the spectrum is more evenly spread, indicating many independent directions.

The \emph{G-Vendi score} \citep{jung2025prismatic} extends this idea by measuring diversity in \emph{gradient space} rather than in representation space. Instead of comparing input examples directly, G-Vendi compares the gradients that each example induces during training. Formally, let $\nabla \ell(x; \theta)$ denote the gradient of the loss with respect to model parameters $\theta$ for a data sample $x$, computed via backpropagation under a proxy model. One benefit of the G-Vendi score is that it encourages orthogonality among representation vectors, thereby promoting diversity in the learned set. In gradient space, this reduces redundancy among update directions, leading to more independent and informative optimization steps. As a result, parameter updates interfere less with one another and yield higher information gain per iteration, improving data efficiency. 



When using cosine similarity, the similarity between two samples is given by the dot product of their $\ell_2$-normalized representations. Let $g_i \in \mathbb{R}^d$ denote the per-sample gradient for data point $i$, and define $X \in \mathbb{R}^{n \times d}$ as the matrix whose $i$-th row is $g_i^\top$, where each $g_i$ is $\ell_2$-normalized. The resulting similarity (kernel) matrix is then given by $K = X X^\top \in \mathbb{R}^{n \times n}$
so that each entry $K_{ij} = g_i^\top g_j$ corresponds to the cosine similarity between gradients of data points $i$ and $j$. Let $\{\sigma_i\}$ denote the eigenvalues of $K$, and define normalized eigenvalues $\lambda_i = \sigma_i / \mathrm{Tr}(K)$, which form a probability distribution. The G-Vendi score is then defined as the exponential of the Shannon entropy of this spectrum:
\begin{equation}
\mathrm{G\text{-}Vendi}(K) = \exp\left(- \sum_i \lambda_i \log \lambda_i \right).
\end{equation}

In practice, computing the full kernel matrix is unnecessary. We exploit the fact that the Vendi score kernel is positive semidefinite and that the Gram matrix shares the same non-zero eigenvalues as the kernel matrix. This allows us to work directly with the Gram matrix in $\mathbb{R}^{d \times d}$, which is significantly more computationally efficient and scalable.

\section{\textsc{Spokes}: optimizing for data diversity in gradient space}

\subsection{\textsc{Spokes}: Scalable optimization to achieve high G-Vendi subsets}
\label{sec:spokes_opt_algo}

Given the strong downstream performance associated with G-Vendi, we introduce \textsc{Spokes}, a method for extracting high G-Vendi subsets from large-scale pretraining corpora. In addition to diversity, our formulation incorporates per-example quality scores, motivated by the role of quality filtering in modern data selection pipelines. We control the trade-off between quality and diversity using a tunable parameter $\alpha \in [0,1]$, where $\alpha = 0$ recovers diversity-only optimization. In our experiments, quality scores are obtained using the FineWeb-Edu classifier \citep{penedo2024fineweb}, though the formulation is compatible with any per-example quality metric.

We begin with a discrete subset selection problem in which the objective is to select a subset $S \subseteq [n]$ of fixed size $k$ that jointly maximizes quality and diversity. Let each example $x_i$ have an associated quality score $q_i$, and let $K_S$ denote the similarity kernel restricted to the selected subset. Following the log-form quality-weighted Vendi objective of \citet{nguyen2024quality}, we define the optimization problem as
\begin{equation}
\label{eq:quality_weighted_vendi_loss}
\max_{S \subseteq [n],\, |S|=k} \quad 
\alpha \ln\left(\frac{1}{k}\sum_{i \in S} q_i\right)
+
(1-\alpha)\ln \mathrm{Vendi}(K_S).
\end{equation}

While this formulation directly captures the desired trade-off between quality and diversity, optimizing it is computationally intractable due to the combinatorial nature of subset selection. As such, we used a relaxed optimization as we describe below.

To construct the similarity representation used by G-Vendi, we first compute gradient-based embeddings using a proxy model:
\begin{equation}
g_i = \nabla_{\theta} \ell(x_i), \quad i=1,\dots,n.
\end{equation}
Because these gradients are high-dimensional, we apply a random projection using a Rademacher matrix $R \in \{-1,+1\}^{k \times d}$:
\begin{equation}
z_i = \frac{1}{\sqrt{k}} R g_i.
\end{equation}

Using the projected embeddings, we define pairwise similarities through a cosine similarity kernel. We then relax the discrete optimization problem by adding a continuous non-negative weight to every data point $w \in \Delta^n$. Under this relaxation, the quality objective becomes the expected quality
\begin{equation}
Q(w) = \sum_{i=1}^n w_i q_i.
\end{equation}

To extend the subset kernel $K_S$ to the relaxed setting, we define a weight-dependent kernel
\begin{equation}
K(w)_{ij}
=
\sqrt{w_i w_j}
\cdot
\frac{z_i^\top z_j}
{\|z_i\| \|z_j\|},
\end{equation}
The resulting relaxed optimization problem is
\begin{equation}
\max_{w \in \Delta^n} \;
\alpha \ln Q(w)
+ (1 - \alpha)\, \ln \operatorname{Vendi}\!\big(K_w\big)
\end{equation}

We optimize this objective using exponentiated gradient descent while enforcing the simplex constraint through normalization at each iteration, preventing collapse onto a small number of high-weight examples. After optimization, we recover a discrete subset by selecting the top-$k$ entries of the learned weight vector. We demonstrate the effectiveness of this strategy in \Cref{fig:gvenditopk}. The complete formulation of \textsc{Spokes} is summarized in \Cref{alg:spokes}.

\begin{algorithm}[h]
\caption{\textsc{Spokes}: Probabilistic G-Vendi via Exponentiated Gradient Descent}
\label{alg:spokes}
\begin{algorithmic}[1]

\STATE {\bf Input:} Dataset $\{x_1, \dots, x_n\}$, quality scores $\{q_i\}_{i=1}^n$, learning rate $\eta > 0$, iterations $T$, initial distribution $w^{(0)} \in \Delta^n$, trade-off parameter $\alpha \in [0,1]$

\STATE {\bf Step 1:} Compute gradient embeddings using a proxy model:
\[
g_i = \nabla_{\theta} \ell(x_i), \quad i=1,\dots,n
\]

\STATE {\bf Step 2:} Sample a Rademacher projection matrix $R \in \{-1,+1\}^{k \times d}$ and project embeddings:
\[
\tilde{z}_i = \frac{1}{\sqrt{k}} R g_i, \quad \forall i
\]

\STATE {\bf Step 3:} Construct weighted cosine similarity kernel:
\[
K(w)_{ij} = \sqrt{w_i w_j} \cdot \frac{z_i^\top z_j}{\|z_i\| \|z_j\|},
\]

\STATE {\bf Step 4:} Define weighted quality:
\[
Q(w) = \sum_{i=1}^n w_i q_i
\]

\STATE {\bf Optimization objective:}
\[
\begin{aligned}
\text{maximize} \quad & \alpha \ln Q(w) + (1-\alpha)\ln \mathrm{Vendi}(K) \\
\text{subject to} \quad & w \in \Delta^n
\end{aligned}
\]

\STATE {\bf Step 5:} Define loss:
\[
\mathcal{L}(w) = -\left( \alpha \ln Q(w) + (1-\alpha)\ln \mathrm{Vendi}(K) \right)
\]

\FOR{$t = 0$ to $T-1$}

    \STATE Compute gradient:
    \[
    g^{(t)} = \nabla_{w} \mathcal{L}(w^{(t)})
    \]

    \STATE Exponentiated gradient update:
    \[
    \tilde{w}_i^{(t+1)} = w_i^{(t)} \exp(-\eta g_i^{(t)}), \quad \forall i
    \]

    \STATE Normalize:
    \[
    w^{(t+1)} = \frac{\tilde{w}^{(t+1)}}{\sum_{j=1}^n \tilde{w}_j^{(t+1)}}
    \]

\ENDFOR

\STATE {\bf Output:} Subset $S = \mathrm{Top\text{-}k}(w^{(T)})$

\end{algorithmic}
\end{algorithm}

\subsection{The optimisation leads to smooth trade-offs between quality and diversity}

\begin{figure}[h]
    \centering
    \begin{minipage}{0.394\textwidth}
        \centering
        \includegraphics[width=\linewidth]{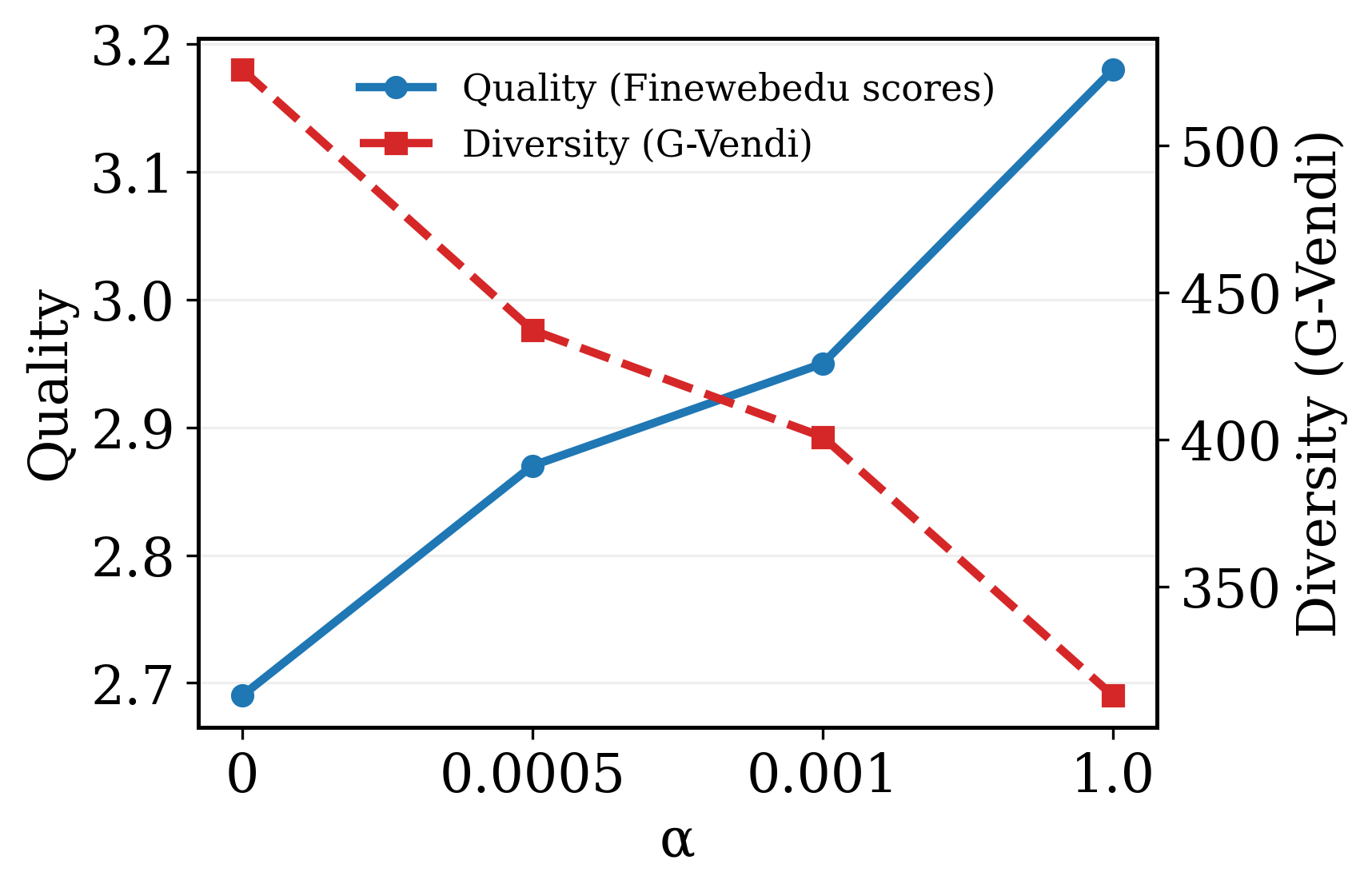}
        \caption{Quality–diversity trade-off across $\alpha$.}
        \label{fig:quality-diversity-tradeoff}
    \end{minipage}
    \hfill
    \begin{minipage}{0.526\textwidth}
        \centering
        \includegraphics[width=\linewidth]{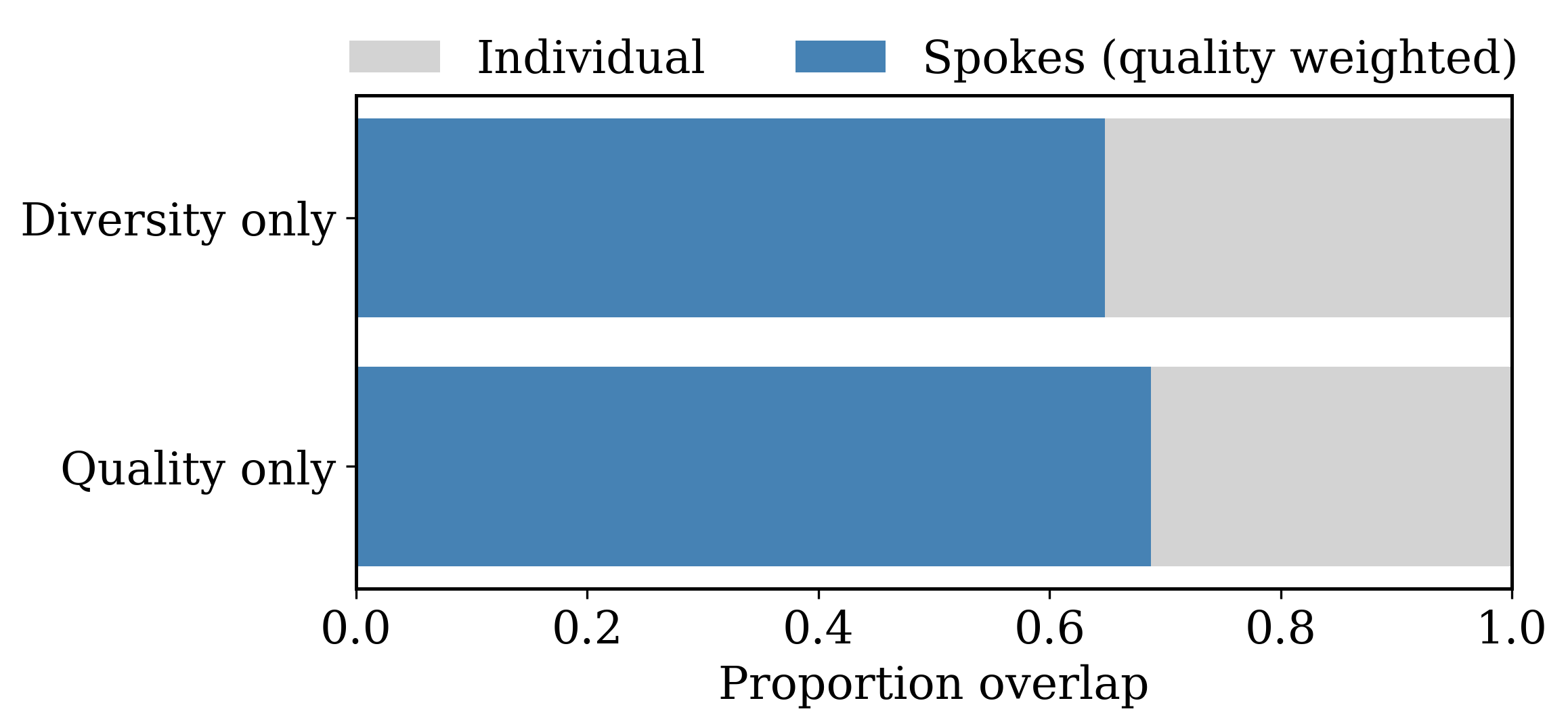}
        \caption{Overlap between quality-only, diversity-only, and joint optimization subsets.}
        \label{fig:qVs-proportion}
    \end{minipage}
\end{figure}

\textsc{Spokes} is effective in practice and we observe a smooth trade-off between quality and diversity. \Cref{fig:quality-diversity-tradeoff} shows that increasing $\alpha$ improves average quality while reducing G-Vendi Score in a controlled manner. This behavior is consistent when we optimise over the full pretraining set and measure the G-Vendi and quality score of a batch sampled from this set. 

To select a final value of $\alpha$, we measure overlap between the jointly optimized subset and the two extreme solutions (quality-only and diversity-only). For DCLM, $\alpha = 0.001$ yields a balanced trade-off, with 64.8\% overlap with the diversity-only subset and 68.8\% overlap with the quality-only subset. For FineWeb, which contains a higher proportion of low-quality documents and therefore induces a natural bias toward higher-quality selections, we use a lower $\alpha = 0.0005$, obtaining 70.0\% and 87.0\% overlap, respectively. In both cases, the chosen $\alpha$ substantially increases the average batch G-Vendi scores relative to quality-only selection.

\subsection{Time complexity of \textsc{Spokes}}
\label{sec:time_complexity}

\textsc{Spokes} can be used as a drop-in alternative to existing data selection methods such as SemDeDup \citep{abbas2023semdedup}. The key advantage of \textsc{Spokes} is improved computational efficiency, achieved by operating directly on a Gram matrix rather than relying on pairwise similarity computations.

At each iteration, \textsc{Spokes} constructs a Gram matrix over the currently selected subset. Let the subset size be \(N\) and the representation dimension be \(d\). The per-iteration computational cost is \(O(N d^2)\), leading to a total cost of \(O(T N d^2)\) over \(T\) optimization steps. In practice, \(T\) is small (typically \(T < 20\)) because the selection objective converges quickly to a representative subset.

To further improve scalability, the dataset can be partitioned into \(p\) disjoint subsets of size approximately \(N/p\), which are optimized independently. This yields a per-partition cost of \(O\!\left(T \frac{N}{p} d^2\right)\), resulting in an approximately linear speedup in \(p\) under the assumption of independent optimization. In our experiments, we set \(p = 1\), as the optimization remains efficient at full scale. For the optimization process, around 8 NVIDIA H100 GPU hours was used.

In contrast, SemDeDup \citep{abbas2023semdedup} relies on pairwise similarity computations between samples, incurring a dominant cost that scales quadratically with the subset size, \(O(N^2 / k)\) where \(k\) is the number of clusters. This quadratic dependence makes SemDeDup substantially more expensive at scale. Moreover, SemDeDup typically requires a larger number of clusters \(k\) than the number of partitions \(p\) used in \textsc{Spokes}, further increasing its computational cost. In our experiments, the semantic deduplication procedure took around 16 NVIDIA H100 GPU hours. 

Beyond efficiency, \textsc{Spokes} also offers greater control over the final subset size. Target dataset sizes can be obtained directly by selecting the top-k samples, whereas SemDeDup requires sweeping over multiple $\epsilon$ thresholds to achieve a desired data budget.

\section{Improving efficiency in \textsc{Spokes}}
Beyond the relaxed optimisation procedure in \Cref{sec:spokes_opt_algo} we perform some modifications to the pipeline to make the optimisation procedure for G-vendi more scalable.

\label{sec:scaling_spokes}
\subsection{Approximating gradients using the last $n$ layers}

Embedding-based approaches, such as those used in semantic deduplication, are already standard practice in pretraining pipelines \citep{abbas2023semdedup}. These methods typically rely on full-model embeddings, which only require one forward pass. In contrast, our method requires gradient-based representations, which are substantially more expensive if computed over the entire model.

To reduce this overhead, we approximate full gradients by restricting computation to the last $n$ transformer layers. This truncation significantly lowers computational cost while preserving the structure of the resulting similarity kernel (see \Cref{tab:grad_corr}). In practice, this makes gradient computation comparable to embedding-based methods when $n$ is small (e.g., 2–3 layers).

To select an appropriate truncation depth, we perform a small calibration study using 100 samples. We compute similarity kernels from truncated gradients and measure their agreement with full-gradient kernels using average row-wise Spearman correlation. Empirically, we find strong agreement between kernel matrices constructed from truncated gradients and those from full gradients. In particular, Spearman correlation between pairwise similarities remains high, with diminishing returns beyond two layers for \texttt{Qwen3-0.6B-Base} and three layers for \texttt{Qwen2.5-0.5B-Instruct}.

\begin{table}[h]
\centering
\caption{Average Spearman correlation with full gradients}
\label{tab:grad_corr}
\begin{tabular}{l|c|c}
\hline
\textbf{N-last layers} & \textbf{Qwen3-0.6B-Base} & \textbf{Qwen2.5-0.5B-Instruct} \\
\hline
Last 2 layers & 0.9308 & 0.2612 \\
Last 3 layers & 0.9310 & 0.9232 \\
Last 4 layers & 0.9312 & 0.9236 \\

\end{tabular}
\end{table}

Since our focus is on pretraining corpora rather than instruction-tuned settings, we use \texttt{Qwen3-0.6B-Base} as the default model for gradient computation and restrict gradient computation to the last two layers, which provides a strong accuracy and efficiency trade-off. 

\subsection{For subset selection a smaller k can be chosen for Johnson--Lindenstrauss projection dimensions}

Despite restricting gradient computation to the final two layers, the resulting gradient representations remain extremely high-dimensional, comprising 200 million parameters. Direct pairwise similarity computation in this space is therefore computationally prohibitive. To address this, we employ a dimensionality reduction scheme based on the Johnson--Lindenstrauss (JL) lemma \citep{johnson1984extensions}, instantiated via Rademacher random projections \citep{achlioptas2003database}, which preserve pairwise distances in low dimensions under projection.

Formally, let $X \subset \mathbb{R}^d$ be a set of $n$ points and let $\varepsilon \in (0,1)$. We construct a random projection matrix $R \in \mathbb{R}^{k \times d}$, where $R_{ij} \in \{+1,-1\}$ are i.i.d. Rademacher variables. If
\begin{equation}
k \ge \frac{4 + 2\beta}{\varepsilon^2/2 - \varepsilon^3/3} \ln(n),
\end{equation}
then with probability at least $1 - n^{-\beta}$, the embedding $f(u) = \frac{1}{\sqrt{k}}Ru$ approximately preserves all pairwise Euclidean distances over $X$, such that
\begin{equation}
(1 - \varepsilon)\|u - v\|^2 \le \|f(u) - f(v)\|^2 \le (1 + \varepsilon)\|u - v\|^2
\quad \forall\, u,v \in X.
\end{equation}

For example, if N = $3 \times 10^8$ and $\varepsilon = 0.11$ and $\beta=1$ this bound yields $k \approx 18035$, which is computationally expensive in both storage and runtime at pretraining scale.

In reality, we are doing subset selection, so even if the similarities are distorted, subsets chosen may still be accurate based on the proportion of data chosen. To study this, we empirically evaluate the effect of $k$ on subset stability. We sample 200k documents and compute gradient-based representations using the last two model layers. We then perform G-Vendi optimization across different values of $k$ and measure the overlap between selected subsets. We sweep $k$ from 512 to 16{,}384 and compare each resulting subset against a high-dimensional reference using $k = 17{,}920$. Despite large changes in projection dimension, subset selection remains highly stable across all settings (See Appendix~\ref{app:k-johnson}).

In particular, even at $k = 1024$, we observe 95.5\% overlap with the reference subset at a 0.5 selection proportion, indicating that low-dimensional JL embeddings are sufficient for G-Vendi optimization in practice. Given the high overlap of subset chosen and the benefits of saving memory with lower dimensions, we chose to use $k=1024$.

\subsection{Controlling for length}

Prior work has shown that sequence length can influence embedding similarity-based metrics. As a diagnostic baseline, we first observe that in an embedding-based Vendi setup using Qwen3-0.6B-embedding model \citep{zhang2025qwen3}, the selected subset has a lower average sequence length (827 tokens) than the random baseline (1230 tokens), despite higher diversity scores. This effect is consistent with known pooling behavior in embedding models, where token representations are aggregated over context.

To reduce length-driven confounding effects in this baseline comparison, we truncate all sequences to a maximum length of 768 tokens, approximately matching the median length of a random DCLM sample (710 tokens). Under this setting, gradient-based subsets still exhibit higher average sequence length (1482 tokens) compared to both embedding-based subsets (1136) and random sampling (1230). However, it is likely that the increased sequence length in the G-Vendi case comes from the natural diversity of the dataset as all samples are subject to the same length constraint. 

\section{Experimental Setup}
\label{sec:experiment}

\subsection{Dataset Selection}

We evaluate our method on two large-scale pretraining corpora with differing filtering regimes to assess robustness across data distributions. 

First, we use the DCLM dataset (Dolmino subset) \citep{olmo20242}, which undergoes relatively aggressive filtering and curation. Second, we use FineWeb \citep{penedo2024fineweb}, a less aggressively filtered web-scale corpus. These datasets differ in their noise profiles and redundancy, providing a useful testbed for evaluating the effect of diversity-based selection under varying data quality conditions.

\subsection{Baselines}

We compare against commonly used data selection strategies in large-scale language model pretraining: (1) \textbf{Random sampling}, which serves as a standard baseline. (2) \textbf{Semantic deduplication}, which removes near-duplicate samples using embedding-based clustering. (3) \textbf{Quality filtering}, which selects high-quality subsets based on heuristic or learned quality scores. The quality scores were based on fineweb-edu classifiers \citep{penedo2024fineweb}.

We evaluate these methods under a pretraining from scratch training regime:  (i) pretraining from scratch on a LLaMA-1B architecture \citep{grattafiori2024llama} on 100 billion tokens.

\subsection{Hyperparameters used}
We used a constant learning rate of 3e-4 for all experiments. We also used a batch size of 2048 with sequence length 4096. All pretraining from scratch experiments was run on a LLaMA 1b architecture using the megatron-lm  library \citep{shoeybi2019megatron}. Each single run took around 576 NVIDIA H100 gpu hours. 

For the semantic deduplication experiments, we tuned the epsilon until we were able to filter 50\% of the data. 10000 clusters were used and an epsilon of 0.005 was used to filter the DCLM subset. For fineweb, a total of 20000 clusters and an epsilon of 0.25 was used. More clusters were used as fineweb has almost twice the number of data samples despite having the same token count. For the quality filters, a threshold of 3.0625 and 1.0693 were used for DCLM and Fineweb respectively.

\subsection{Evaluation}
We evaluate models on 10 English benchmarks from the OLMES evaluation suite \citep{gu2025olmes}, which was chosen for its strong discriminative power among small models and its use of standardized prompting, consistent formatting, and cloze-style task formulations.

\section{Results}

\subsection{\textsc{Spokes} improves batch level gradient diversity}

\begin{figure}[h]
    \centering
    \begin{minipage}{0.5075\textwidth}
        \centering
        \includegraphics[width=\linewidth]{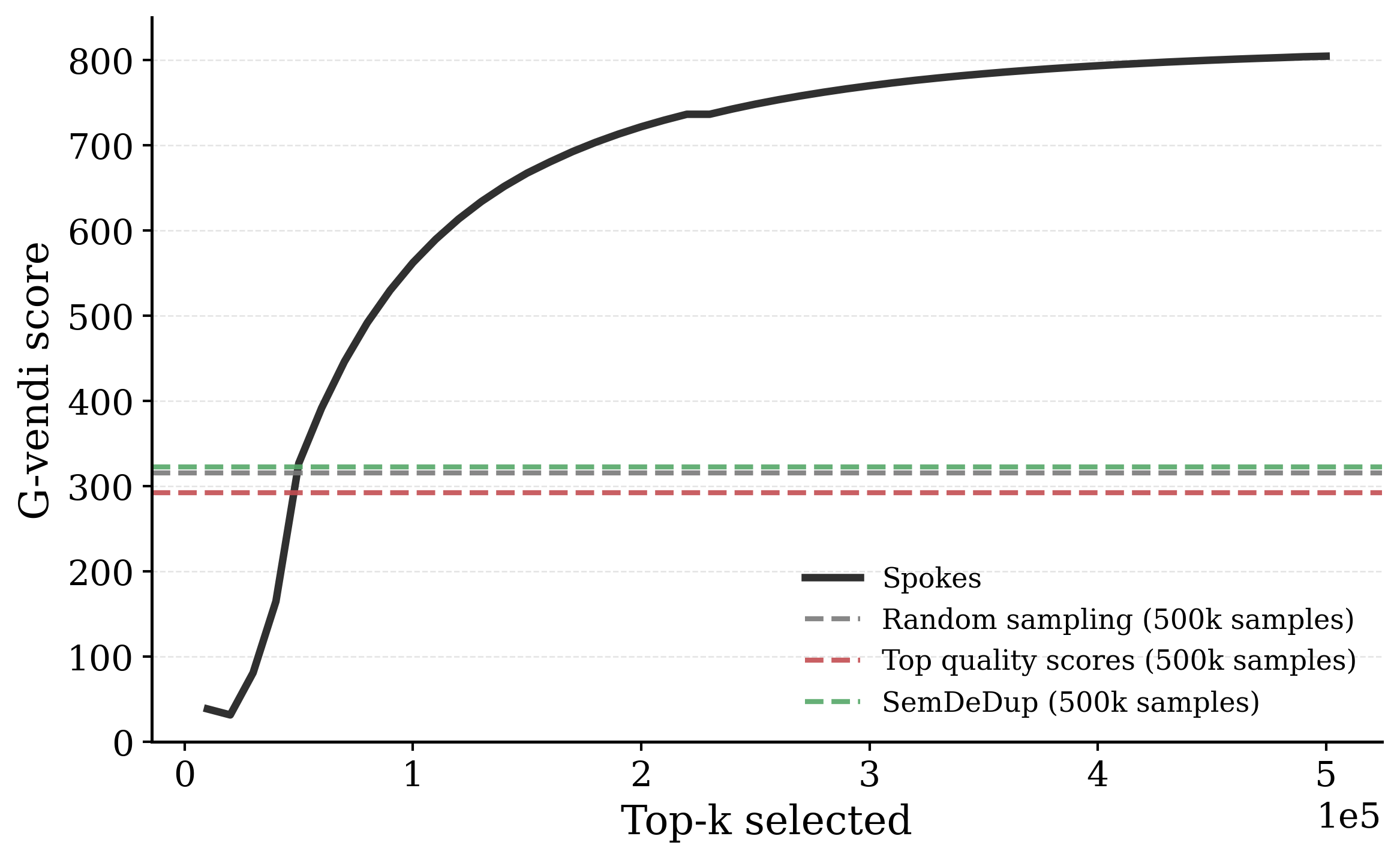}
        \caption{G-vendi scores scale as the number of top-k increases}
        \label{fig:gvenditopk}
    \end{minipage}
    \hfill
    \begin{minipage}{0.410\textwidth}
        \centering
        \includegraphics[width=\linewidth]{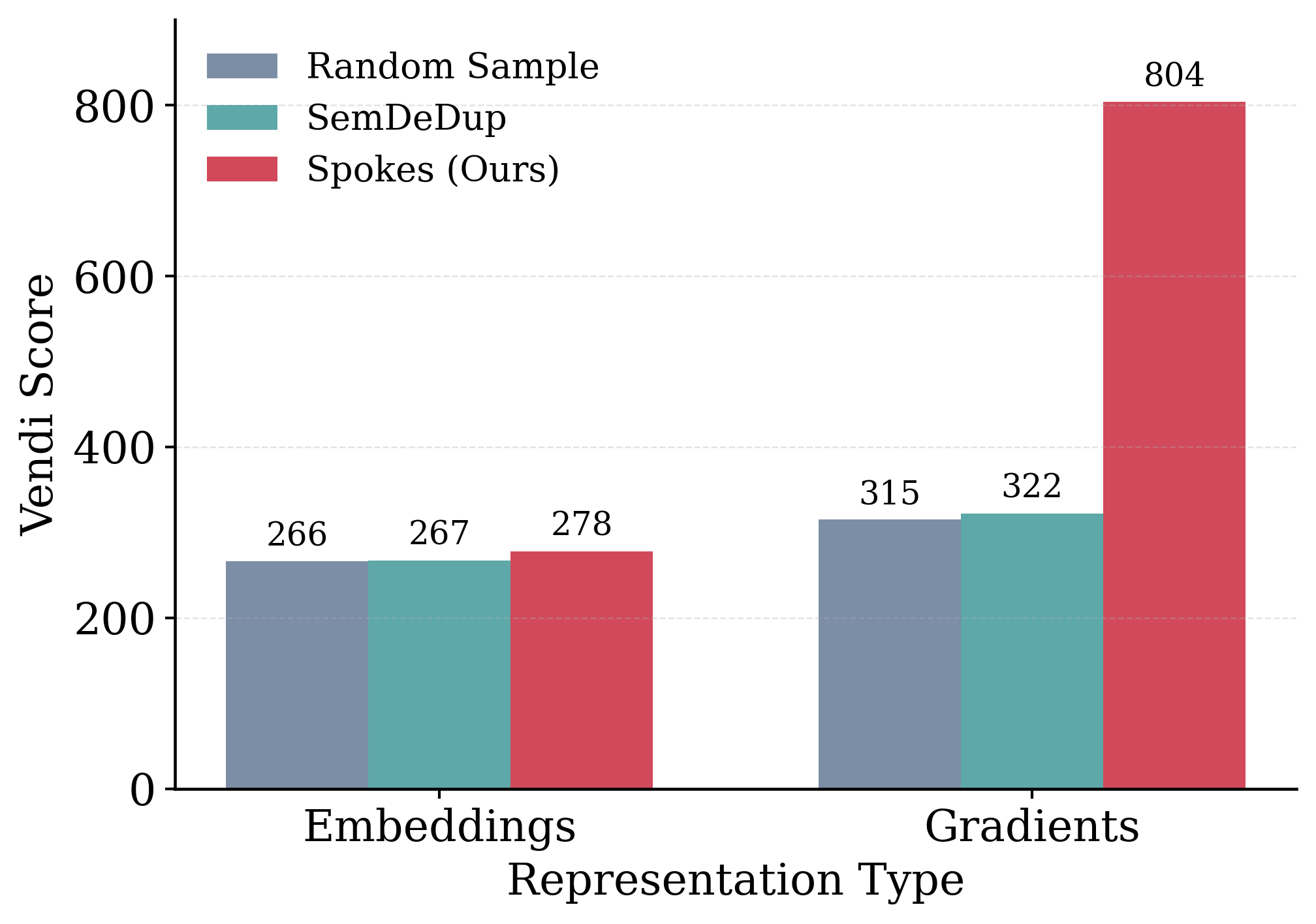}
        \caption{Vendi score using different representations and optimisation techniques}
        \label{fig:gvendi_rep}
    \end{minipage}
\end{figure}

We find that our optimization procedure is able to extract substantially more diverse subsets from pretraining corpora than are typically recovered by random sampling. For example, a random sample of 500k documents drawn from a 200B-token pool of DCLM \citep{li2024datacomp} from the Dolmino dataset \citep{olmo20242} (173M documents), achieves a relatively low G-Vendi score of 315, as shown in \Cref{fig:gvendi_rep}. In contrast, using the optimised weights of \textsc{Spokes} to select k data points  produces a consistent scaling effect: as top-k based on the weights optimized by \Cref{alg:spokes} increases, the G-Vendi score improves monotonically, as shown in \Cref{fig:gvenditopk}. This shows that \textsc{Spokes} is highly effective in recovering high diversity subsets, which was unable to be captured by uniform sampling. 

The benefits of our subset selection extend beyond identifying small, diverse sets at the highest weight points. When performing full subset selection for the experiments in \Cref{sec:experiment}, we observe that the average batch diversity remains consistently high throughout training, increasing from 315 to 520. This indicates that higher-diversity batches are encountered at every update step of the pretraining stage with the diverse selected data.

\subsection{Gradients are more amenable to optimization}
We compare different representations for measuring diversity, focusing on embeddings and gradients. As shown in \Cref{fig:gvendi_rep}, embedding-based representations saturate quickly on large corpora, limiting their ability to discriminate between subsets. In contrast, gradient-based representations remain sensitive under optimization.

While random subsets achieve similar Vendi scores under both representations, applying our selection procedure leads to a substantial increase in G-Vendi score when using gradients, whereas embeddings show limited improvement. This indicates that gradients capture finer-grained structure that is not reflected in embedding similarity alone. This observation is consistent with prior work showing stronger alignment between gradient-space structure and out-of-distribution generalization \citep{jung2025prismatic}, which justifies our shift to gradients rather than embeddings.
\subsection{Spokes improves pretraining outcomes reliably across different datasets}

\begin{table}[h]
\centering
\caption{Performance on world knowledge benchmarks. Best results are \best{highlighted}, second best are \second{underlined}.}
\label{tab:pt_results}
\normalsize
\resizebox{\textwidth}{!}{
\begin{tabular}{l|cccccccccc|c}
\hline
\textbf{Corpus} 
& \textbf{HellaSwag} 
& \textbf{MMLU} 
& \textbf{ARC-C} 
& \textbf{ARC-E} 
& \textbf{PIQA} 
& \textbf{SIQA} 
& \textbf{BoolQ} 
& \textbf{WinoGrande} 
& \textbf{OBQA} 
& \textbf{CSQA}
& \textbf{Avg} \\ 

\hline
\multicolumn{12}{l}{\textit{DCLM}} \\

DCLM (Random) 
& 46.1 & 34.0 & 39.8 & \second{72.6} & \best{69.3} & 46.3 & 63.5 & 53.0 & 45.0 & 55.5 & 52.5 \\

DCLM (SemDeDup) 
& 46.4 & 34.3 & 41.8 & 70.0 & 67.8 & 46.3 & \second{63.8} & 53.2 & 43.4 & \second{57.9} & 52.5 \\

\rowcolor{rowgray}
DCLM-\textsc{Spokes} (Ours) 
& \best{49.8} & 34.3 & 41.3 & 72.0 & \best{69.3} & \best{46.9} & 62.6 & 53.4 & \second{46.2} & 53.1 & 52.9 \\

DCLM-Quality (Top 50\%) 
& 46.6 & \best{35.1} & \best{43.5} & \best{74.2} & 66.7 & 45.0 & \second{63.8} & \best{54.5} & 46.0 & 54.7 & \second{53.0} \\

\rowcolor{rowgray}
DCLM-\textsc{Spokes} (Quality-weighted, Ours) 
& \second{48.7} & \best{35.1} & \second{42.9} & \second{72.6} & 67.7 & \best{46.9} & \best{66.3} & \best{54.5} & \best{47.0} & \best{58.6} & \best{54.0} \\
\hline

\multicolumn{12}{l}{\textit{FineWeb}} \\

FineWeb (Random) 
& 46.4 & 27.8 & \second{29.4} & \second{52.4} & 66.9 & 41.5 & \best{64.1} & \second{52.2} & 32.8 & 37.2 & 45.1 \\

FineWeb (SemDeDup) 
& 45.0 & 27.9 & 27.0 & 49.2 & 65.7 & 40.6 &  \underline{63.8} & 51.7 & \best{37.8} & 31.8 & 44.0 \\

\rowcolor{rowgray}
FineWeb-\textsc{Spokes} (Ours) 
& \best{48.5} & \best{29.4} & \second{29.4} & 51.9 & 67.4 & 40.8 & 57.5 & 51.4 & \second{36.0} & \best{44.1} & \underline{45.6} \\

FineWeb-Quality (Top 50\%) 
& \best{48.5} & 28.4 & \second{29.4} & 48.4 & \second{69.3} & \best{42.4} & 49.5 & \best{54.3} & 35.0 & \second{40.8} & 44.6 \\

\rowcolor{rowgray}
FineWeb-\textsc{Spokes} (Quality-weighted, Ours) 
& 47.9 & \second{28.7} & \best{30.9} & \best{54.7} & \best{69.4} & \best{42.4} & 63.7 & 51.7 & \second{36.0} & 39.9 & \best{46.5} \\

\hline
\end{tabular}
}
\end{table}

Overall, our results show that: (1) Diversity optimization alone yields consistent improvements over random sampling, (2) Jointly optimizing quality and diversity produces the best overall performance, outperforming all other baselines. (3) Quality and diversity play a complementary role to each other.

\paragraph{Diversity alone provides a strong signal.}
Across both DCLM and FineWeb, \textsc{Spokes} consistently outperforms random sampling. On DCLM, diversity-only selection improves the average score from 52.5 to 52.9; on FineWeb, from 45.1 to 45.6.
These results show that optimizing for diversity alone yields consistent gains across corpora, even without explicit quality supervision. This suggests that a more diverse gradient representation of the underlying data distribution is sufficient to improve downstream generalization.

\paragraph{Joint optimization yields the best overall performance.}
Using joint optimiziation of both quality and diversity with \Cref{alg:spokes}, we were able to achieve the strongest overall results that outperform both the quality-only and diversity-only baselines. The gains were much more significant, improving from 52.5 to 54.0 for DCLM and 45.1 to 46.5 in Fineweb. 

Notably, jointly optimized subsets have \emph{lower} average quality scores than quality-only subsets (2.95 vs.\ 3.18), yet achieve better downstream performance. This discrepancy suggests that standard quality metrics do not fully capture data utility and including diversity in the optimization can contribute complementary and useful training signals.

\paragraph{Quality and diversity capture complementary strengths.}
Performance differences across tasks reveal a clear trade-off. For DCLM,  Diversity-only selection performs well on commonsense reasoning tasks (e.g., HellaSwag \citep{zellers2019hellaswag}), while quality-based filtering is more effective on knowledge-intensive benchmarks (e.g., MMLU \citep{hendrycks2020measuring} ).
\textsc{Spokes} bridges this gap. Joint optimization matches or exceeds the stronger baseline on most tasks while avoiding the weaknesses of optimizing for either objective alone. For example, on DCLM, diversity improves HellaSwag  but lags on MMLU relative to quality filtering, whereas \textsc{Spokes} remains competitive on both.
These results indicate that quality and diversity encode distinct but complementary properties of the training data that are both useful for subset construction under fixed data budgets.

\section{Limitations}
\label{sec:limitations}
We acknowledge that a key limitation of our method is the computational cost of gradient evaluation. While we restrict computation to the final two layers, which already yields a significant speedup, there remains substantial room for further efficiency improvements. Prior work has shown that gradient computation can be made effectively compute-optimal \citep{yin2024compute} in regimes where the fixed cost of training the model dominates the cost of gradient evaluation. This assumption aligns with our setting where we expect selected data to be reused over multiple large pretraining runs. Additional acceleration strategies may be possible by leveraging modern techniques such as Cut Cross Entropy \citep{wijmans2024cut}, which reduce memory overhead and improve the efficiency of gradient computation, particularly in smaller models.


\section{Conclusion}

We propose \textsc{Spokes}, a scalable method for diverse subset selection using G-vendi scores. Diversity, as a set-level property, has remained challenging to optimize directly, with prior approaches typically relying on proxy objectives or heuristics such as coarse grained clustering. \textsc{Spokes} addresses this gap through a principled, global optimization framework that enables direct, data point level subset selection, producing consistently more diverse subsets than existing baselines.

Across multiple corpora, \textsc{Spokes} yields subsets that improve downstream performance over all baseline methods under matched data budgets. Beyond diversity alone, \textsc{Spokes} supports the joint optimization of quality and diversity, enabling the construction of subsets that better balance complementary training signals. Empirically, we find that subsets selected by  \textsc{Spokes} consistently outperform those produced by prior methods across evaluation benchmarks.

These findings suggest that direct optimization of diversity is not only tractable at scale, but also practically beneficial, positioning \textsc{Spokes} as a useful tool for constructing both high-quality and diverse subsets from large pretraining corpora.

\bibliography{paper}
\bibliographystyle{neurips}








\newpage
\appendix
\section{Finding k for johnson lindenstrauss transform}
\label{app:k-johnson}

We empirically study an appropriate choice of \(k\) for applying a Johnson--Lindenstrauss transform to project high-dimensional gradients into a lower-dimensional space. We evaluate this by measuring how closely the selected subset matches a reference subset obtained in a high-dimensional setting, using overlap as the agreement metric. In our experiments, the reference dimensionality corresponds to \(k = 17{,}920\).

We observe that substantially smaller values of \(k\) are sufficient in practice. For instance, at a 50\% subset selection ratio, using \(k = 1{,}024\) achieves a 95.5\% overlap with the subset selected under the reference dimensionality. Given the scale of our experiments, this level of agreement is considered acceptable.

\begin{figure}[h]
    \centering
    \includegraphics[width=0.8\linewidth]{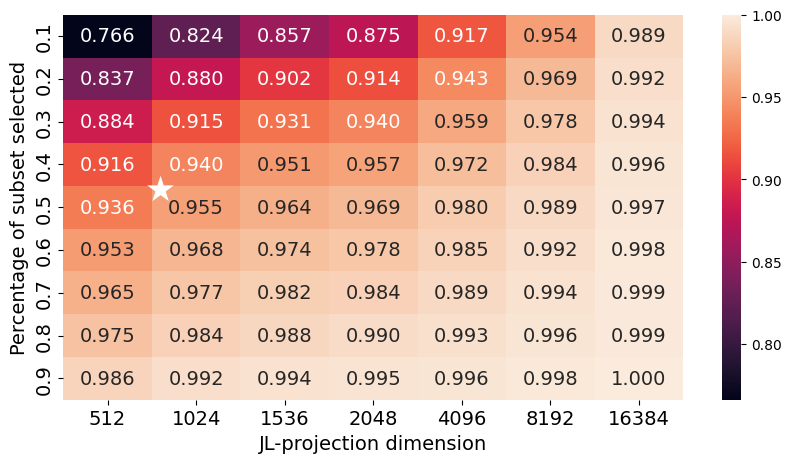}
    \caption{Subset overlap (\%) with $k = 17{,}920$ reference across different Johnson--Lindenstrauss projection dimensions. The white star shows our chosen dimension and subset proportion}.
    \label{fig:heatmap}
\end{figure}

\section{Societal Impact}
\label{app:societal-impact}
Our work advances a more principled understanding of data diversity in LLM pretraining, with potential benefits for improving generalization through training on diverse data. These methods should be applied with careful attention to ethical guidelines and responsible model deployment.



\newpage

\end{document}